\documentclass[conference]{IEEEtran}
\IEEEoverridecommandlockouts
\usepackage{cite}
\usepackage{amsmath,amssymb,amsfonts}
\usepackage{algorithmic}
\usepackage{graphicx}
\usepackage{textcomp}
\usepackage{xcolor}
\usepackage{dsfont}
\def\BibTeX{{\rm B\kern-.05em{\sc i\kern-.025em b}\kern-.08em
    T\kern-.1667em\lower.7ex\hbox{E}\kern-.125emX}}
\begin{document}

\title{HyperTaxa: Multiscale Genome Representations for Generalized Microbial Taxonomic Inference

\thanks{This research was supported in part by the US Department of Agriculture (USDA) grants AP20VSD and B000C011, and the US National Science Foundation grant IIS 2615771.}
}

\author{\IEEEauthorblockN{Akash Rao}
\IEEEauthorblockA{\textit{CSSE Department} \\
\textit{Auburn University}\\
Auburn, AL, USA \\
azr0187@auburn.edu}
\and
\IEEEauthorblockN{Sushim Gupta}
\IEEEauthorblockA{Oklahoma Animal Disease Diagnostic Laboratory \\
\textit{Oklahoma State University}\\
Stillwater, OK, USA \\
sushim.gupta@okstate.edu}
\and
\IEEEauthorblockN{Vishalini Ramnth}
\IEEEauthorblockA{\textit{CSSE Department} \\
\textit{Auburn University}\\
Auburn, AL, USA \\
vlr0013@auburn.edu}
\and
\IEEEauthorblockN{Akhilesh Ramanchandran}
\IEEEauthorblockA{Oklahoma Animal Disease Diagnostic Laboratory \\
\textit{Oklahoma State University}\\
Stillwater, OK, USA \\
rakhile@okstate.edu
}
\and
\IEEEauthorblockN{Sathyanarayanan N. Aakur}
\IEEEauthorblockA{\textit{CSSE Department} \\
\textit{Auburn University}\\
Auburn, AL, USA \\
san0028@auburn.edu}
}

\maketitle

\begin{abstract}
Reliable microbial taxonomic inference is challenged by the rapid growth of newly sequenced organisms and incomplete reference coverage. While alignment- and database-driven tools remain essential for reference-backed identification, they are not designed for learned inference over species absent from genome-level training. We introduce \textsc{HyperTaxa}, a taxonomy-aligned representation learning framework for open-set microbial taxonomic inference. \textsc{HyperTaxa} encodes whole genomes as multiscale hyperspectral $k$-mer transition fields and aligns these representations to a taxonomy-derived semantic space using a two-stage Node2Vec-to-MDS procedure that combines local graph structure with global hierarchical distance preservation. The resulting space supports zero-shot inference over species excluded from genome-level training but represented in the taxonomy, generalized zero-shot inference over seen and unseen species, and cascaded genus-to-species prediction. Across benchmark and large-scale disjoint-species evaluations, \textsc{HyperTaxa} improves taxonomic prediction over pseudo-image and transformer-graph baselines, with the strongest gains at higher taxonomic ranks. Ablations further show that multiscale transition encoding and taxonomy-aware alignment are key contributors to generalization. These results position \textsc{HyperTaxa} as a complementary taxonomy-aware representation and candidate-generation framework for microbial identification under incomplete reference coverage and open-set uncertainty. 
\end{abstract}


\section{Introduction}

Microbial taxonomic inference is fundamental to clinical diagnostics, public health, agriculture, environmental monitoring, and microbial systematics. As sequencing produces increasing numbers of newly assembled isolates and metagenome-assembled genomes, an important computational challenge is to reason about organisms that are rare, under-characterized, or absent from the training distribution. We study this problem as open-set representation learning: given genome-derived sequence evidence, infer taxonomic identity while generalizing to species with no genome examples during training but whose taxonomic placement is known. Zero-shot (ZSL) and generalized zero-shot learning (GZSL) naturally frame this problem by representing unseen species through structured class-level information. Effective inference requires genome representations that capture multiscale sequence dependencies and an embedding geometry that reflects taxonomic organization rather than training-label frequency.

\begin{figure*}
\centering
\includegraphics[width=0.9\linewidth]{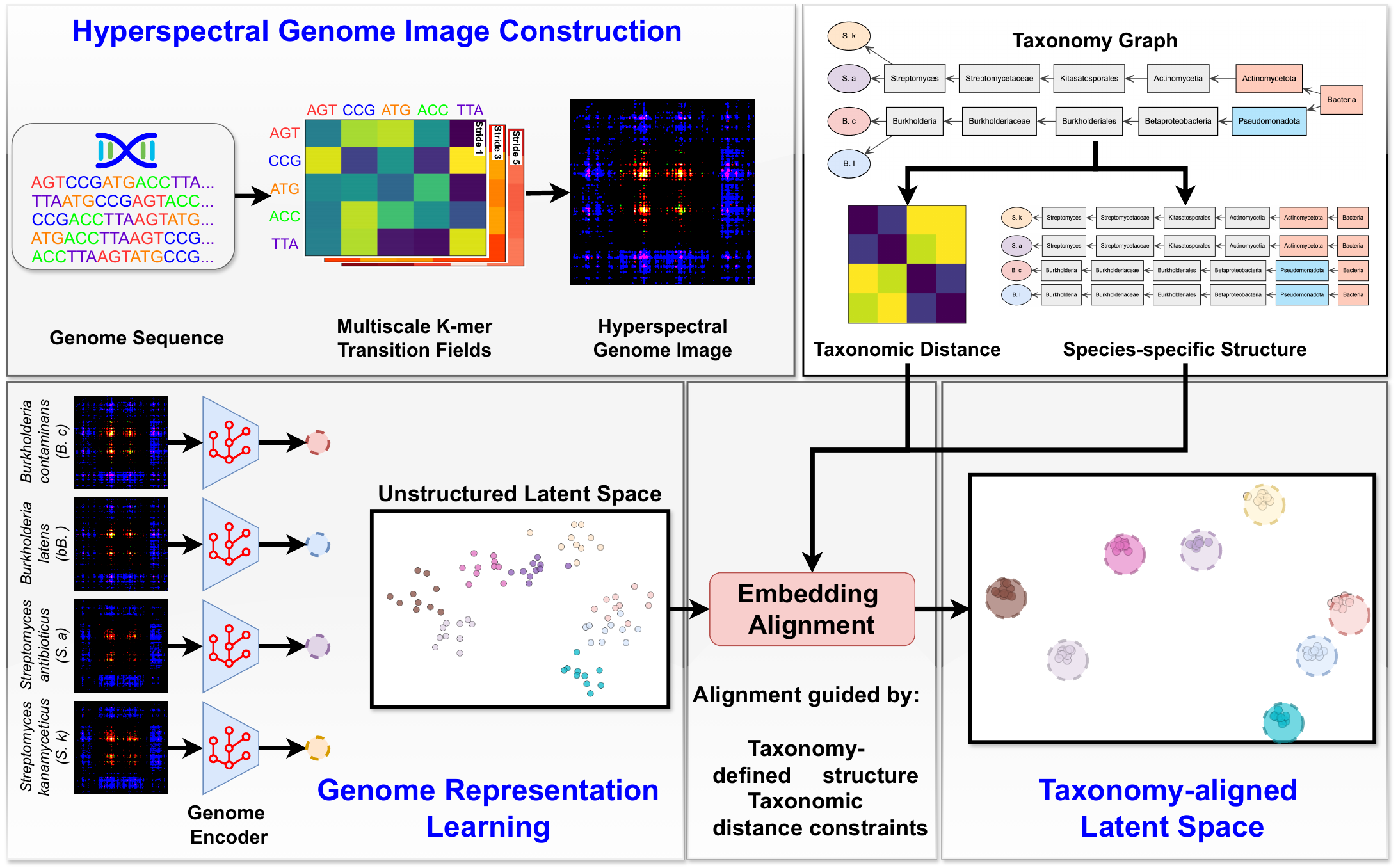}
\caption{\textbf{Overview of \textsc{HyperTaxa}.} Multiscale $k$-mer transition fields encode genome structure, while the microbial taxonomy provides hierarchical and distance constraints. A genome encoder maps these representations into a latent space aligned with taxonomy-derived embeddings, enabling zero-shot and generalized zero-shot inference.}
\label{fig:arch}
\end{figure*}

Existing approaches address related but distinct settings. Reference-based methods such as AAI/ANI \cite{Konstantinidis2005GenomeBasedTaxonomy}, MyTaxa \cite{Luo2014MyTaxa}, EzBioCloud \cite{Yoon2017EzBioCloud}, TYGS \cite{MeierKolthoff2019TYGS}, and GTDB-Tk \cite{Chaumeil2020GTDBTk} remain essential when close references are available, but are not learned zero-shot classifiers over species excluded from training. Supervised genomic models based on codon usage \cite{Khomtchouk2020CodonUsage,Kim2023CodonUsageML,Absci2023DeepCodonLM,Lee2025CodonTransformer}, recurrent networks \cite{Liang2020DeepMicrobes,Shen2018RNN_TFBS,Lanchantin2017DeepMotif}, transformers \cite{Mock2022BERTax}, and metagenomic classification \cite{Mathieu2022MLDLMetagenomics,Park2023MLMarineMetagenomes} likewise remain constrained by observed labels. Species-delimitation \cite{Karbstein2024SpeciesDelimitation4,Perez2025DeepID,Koesters2025DataFusion,Singhal2025GenomicSpeciesDelimitation} and phenotype-prediction methods \cite{Koblitz2025BacterialPhenotypeML,vanHilten2021GenNet,Sehrawat2023NovGMDeep} show that learned representations capture meaningful genomic structure. Mathematical genome mappings \cite{Swidan2006MAGIC,Kaye2021GenomeAtlas,Belokopytova2021PredictingGenomeArchitecture}, graph-based representations and visualization \cite{Gonnella2019GfaViz,Parveen2019CircularVisualization,Grant2023Proksee,narayanan2020gradl}, hyperdimensional encodings \cite{Stock2024Hyperdimensional,He2024HyperGen}, and image-based representations \cite{aakur2024tepi,aakur2021mg} further motivate expressive genome representations. However, these methods do not explicitly align genome features with a taxonomic semantic space for ZSL/GZSL inference.

We introduce \textsc{HyperTaxa}, a taxonomy-aligned framework for open-set microbial taxonomic inference. As shown in Figure~\ref{fig:arch}, genomes are encoded as multiscale hyperspectral $k$-mer transition fields, with channels capturing sequence dependencies at different genomic strides. A convolutional encoder maps these tensors into a latent space aligned with taxonomy-derived embeddings. Node2Vec \cite{grover2016node2vec} captures local graph neighborhoods, while multidimensional scaling (MDS) over taxonomic distances imposes global hierarchical geometry. The resulting space supports ZSL over species unseen during genome-level training, GZSL over seen and unseen species, and cascaded genus-to-species inference. \textsc{HyperTaxa} is complementary to reference-based tools such as BLAST and GTDB-Tk, providing a learned taxonomy-aware representation and candidate-generation layer rather than replacing deterministic sequence comparison.

Our \textbf{contributions} are threefold: (1) we formulate microbial taxonomic inference as ZSL/GZSL over species excluded from genome-level training but represented in the taxonomy; (2) we introduce multiscale hyperspectral $k$-mer transition fields with Node2Vec-to-MDS taxonomy alignment; and (3) we evaluate the framework on the 97-species TEPI benchmark \cite{aakur2024tepi}, a large-scale split with 1{,}110 training and 219 unseen species, and systematic representation and encoder ablations.

\section{Methods}

\textbf{Problem Formulation.}
We formulate microbial taxonomic inference as open-set representation learning over a taxonomic hierarchy. Let $\mathcal{S}_{\mathrm{seen}}$ denote species with genome examples during training and $\mathcal{S}_{\mathrm{unseen}}$ species absent from genome-level training but present in the taxonomy graph. Given a query genome $\mathrm{G}_q$, the goal is to predict its identity at species, genus, family, and order ranks. ZSL evaluates only $\mathcal{S}_{\mathrm{unseen}}$, while GZSL evaluates $\mathcal{S}_{\mathrm{seen}}\cup\mathcal{S}_{\mathrm{unseen}}$.
\begin{figure*}[t]
    \centering
    \begin{tabular}{cccc}
        \includegraphics[width=0.2\linewidth]{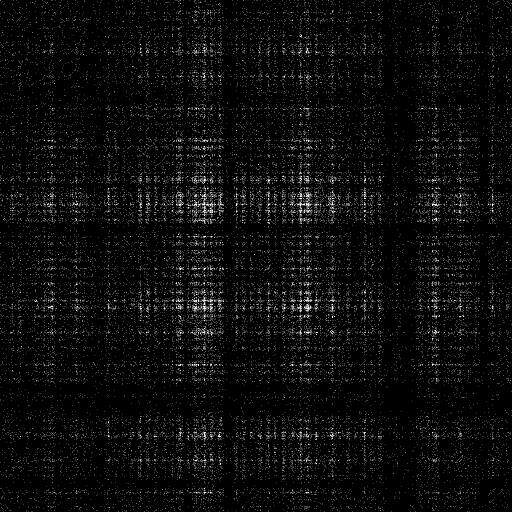} &
        \includegraphics[width=0.2\linewidth]{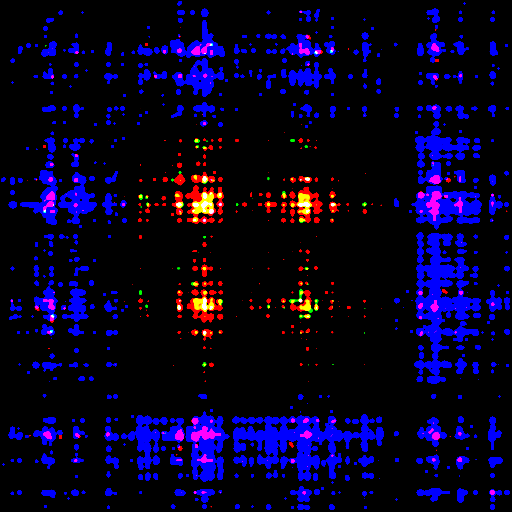} &
        \includegraphics[width=0.22\linewidth]{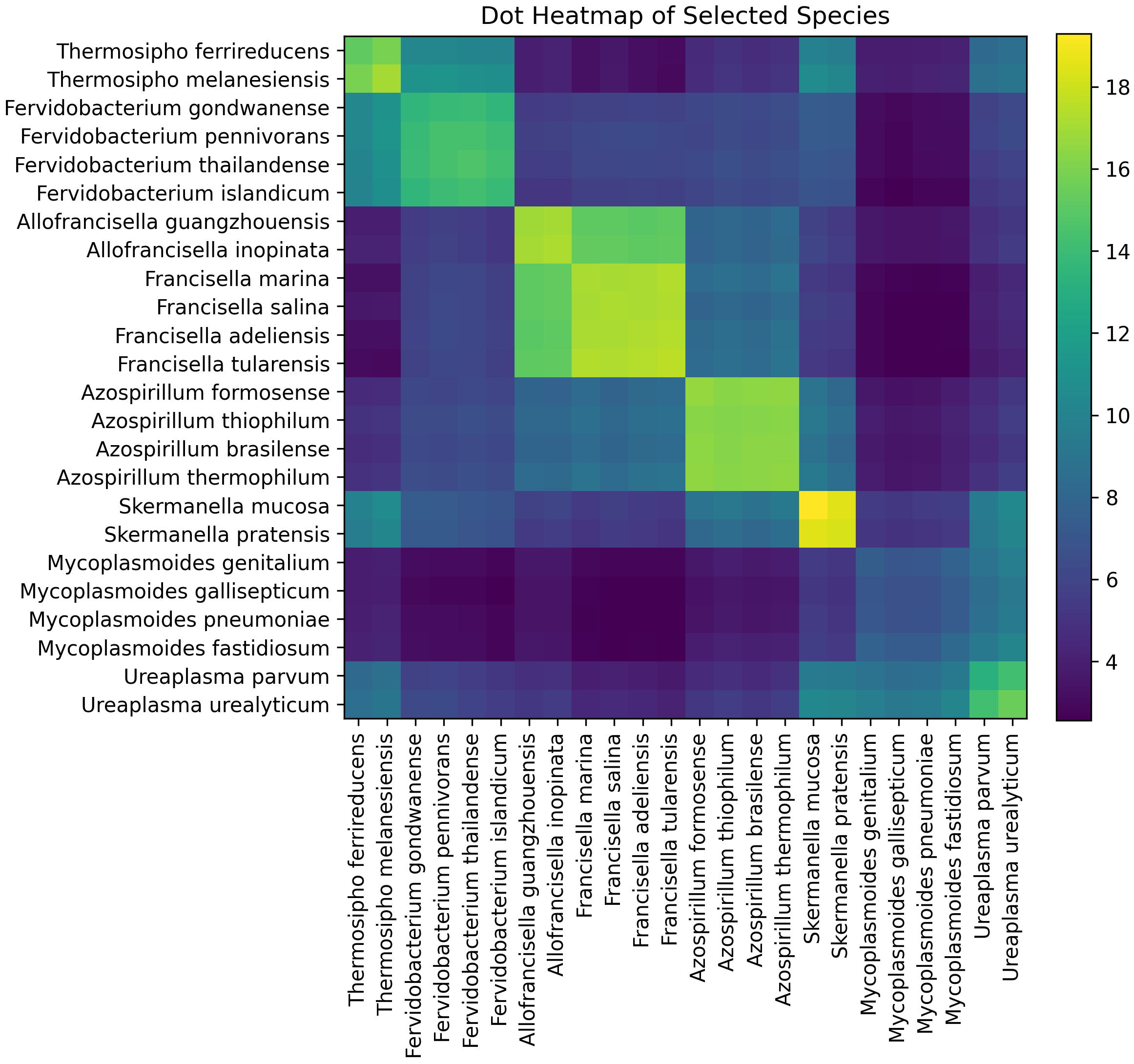} &
        \includegraphics[width=0.22\linewidth]{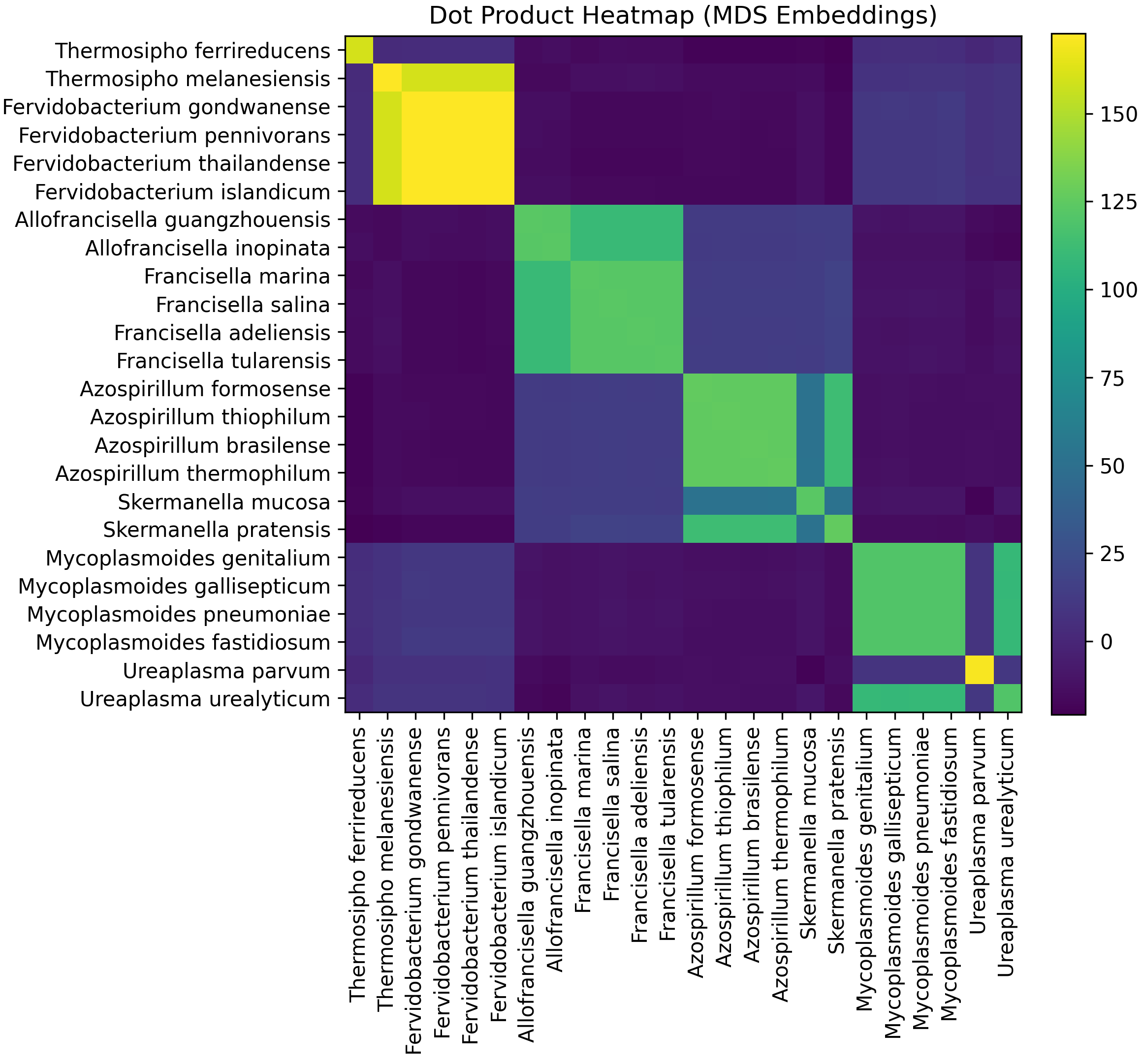} \\
        \textbf{a. TEPI Pseudo-Image} &
        \textbf{b. HyperTaxa Image*} &
        \textbf{c. Node2Vec Similarity} &
        \textbf{d. MDS Similarity}
    \end{tabular}
    \caption{\textbf{Genome representation and taxonomy embedding geometry.}
    \textbf{a--b.} TEPI pseudo-image and \textsc{HyperTaxa} multiscale $k$-mer transition representation for \emph{B. contaminans}.
    \textbf{c--d.} Node2Vec and MDS taxonomy-embedding similarities; MDS yields clearer taxonomic separation. *contrast-enhanced.}
    \label{fig:representation_geometry}
\end{figure*}
Each species $v$ is represented by a taxonomy-derived embedding $\mathbf{e}_v\in\mathbb{R}^d$, including unseen species for which no training genomes are available. A genome encoder $f_\theta$ maps the hyperspectral representation $\mathrm{X}$ to $\mathbf{h}(\mathrm{G})=f_\theta(\mathrm{X})$, and prediction is performed by nearest-neighbor matching to candidate taxon embeddings.

\subsection{Hyperspectral Genome Representation}

To capture multiscale dependencies in whole bacterial genomes, we convert each genome into a \emph{hyperspectral $k$-mer transition tensor}. This representation summarizes how short nucleotide motifs transition into one another when the genome is sampled at different spatial scales, thereby encoding both local sequence composition and longer-range genomic organization. Unlike conventional pseudo-image representations or frequency-based $k$-mer histograms, the proposed representation preserves directional motif-transition structure across multiple genomic strides. 
Formally, let the genome be $\mathrm{G} = (g_1, \dots, g_L)$ with $g_i \in \{A,C,G,T\}$, and fix a $k$-mer size $k$. Let $\mathcal{M} = \{A,C,G,T\}^{k}$ denote the lexicographically ordered set of all $k$-mers, and let $\pi : \mathcal{M} \rightarrow \{1,\dots,4^k\}$ be an indexing map. For a stride $s$, we traverse the genome using hop size $s$, producing sampled positions
$p_n = 1 + (n-1)s$, 
where $p_n + k - 1 \leq L$. The corresponding indexed $k$-mers are 
$u_n^{(s)} = \pi\left(\mathrm{G}[p_n : p_n + k - 1]\right)$. 
Consecutive sampled $k$-mers induce directed transitions $(u_n^{(s)} \rightarrow u_{n+1}^{(s)})$. For each stride $s$, these transitions are accumulated into a matrix $\mathrm{M}^{(s)} \in \mathbb{R}^{4^k \times 4^k}$:
\begin{equation}
\label{eq:transition}
\mathrm{M}^{(s)}_{i,j} =
\sum_{n=1}^{N_s - 1}
\alpha_{n}^{(s)}
\mathds{1}\!\left(u_n^{(s)} = i\right)
\mathds{1}\!\left(u_{n+1}^{(s)} = j\right).
\end{equation}
where $\alpha_{n}^{(s)}$ is a stride-specific transition weight. In practice, each channel is normalized as 
$\widetilde{\mathrm{M}}^{(s)} = \mathrm{M}^{(s)} \Big/ \sum_{i,j}\mathrm{M}^{(s)}_{i,j}$. 
Stacking transition matrices from $C$ strides yields the hyperspectral genome tensor 
$\mathrm{X} =
\big[
\widetilde{\mathrm{M}}^{(s_1)},
\widetilde{\mathrm{M}}^{(s_2)},
\dots,
\widetilde{\mathrm{M}}^{(s_C)}
\big]
\in \mathbb{R}^{4^k \times 4^k \times C}$.
For complete genomes, transition counts are accumulated by traversing the concatenated RefSeq sequence. 
Small strides emphasize local dependencies, including codon usage patterns and short-range compositional motifs, while larger strides capture longer-range structure such as compositional domains and genome-scale organization. Thus, each pixel $(i,j,c)$ measures the directional association between $k$-mer $i$ and $k$-mer $j$ at stride $s_c$. Figure~\ref{fig:representation_geometry}a compares this representation with the TEPI pseudo-image encoding, illustrating that hyperspectral transition fields produce structured multiscale motif patterns rather than sparse positional grids.

\subsection{Taxonomy-Aware Semantic Embedding Construction}

Zero-shot inference requires a class-level embedding for each candidate species, including species with no training genomes. We therefore construct taxonomy-derived embeddings whose geometry reflects both local neighborhood structure and global hierarchical distance in the microbial taxonomy graph. 
Let $\mathcal{V}$ denote the set of species nodes and let $\mathbf{z}_v \in \mathbb{R}^{d}$ denote the desired embedding of species $v \in \mathcal{V}$. The embedding space should satisfy two complementary properties: nearby taxa in the taxonomy graph should have similar embeddings, and Euclidean distances between embeddings should preserve taxonomic distances. We express this objective as
\begin{equation}
\label{eq:tax_geometry}
\mathcal{L}_{\mathrm{tax}}\big({\mathbf{z}_v}\big)
=\lambda_{\mathrm{loc}}
\mathcal{L}_{\mathrm{loc}}\big({\mathbf{z}_v}\big)
+
\lambda_{\mathrm{glob}}
\mathcal{L}_{\mathrm{glob}}\big({\mathbf{z}_v}\big),
\end{equation}
where $\mathcal{L}_{\mathrm{loc}}$ encourages local graph consistency and $\mathcal{L}_{\mathrm{glob}}$ encourages agreement with the shortest-path taxonomic distance $d_{\mathrm{tax}}(u,v)$. 
Rather than directly optimizing Eq.~\eqref{eq:tax_geometry}, we approximate its two terms using two complementary embeddings. First, we compute a Node2Vec embedding $\mathbf{e}^{\mathrm{n2v}}_v \in \mathbb{R}^{d}$ for each species node using random walks on the taxonomy graph. This captures local co-occurrence and neighborhood structure. Second, we compute all-pairs shortest-path distances $d_{\mathrm{tax}}(u,v)$ between species and apply classical multidimensional scaling (MDS) to obtain $\mathbf{e}^{\mathrm{mds}}_v \in \mathbb{R}^{d}$ such that 
$\left|
\mathbf{e}^{\mathrm{mds}}_u -
\mathbf{e}^{\mathrm{mds}}_v
\right|_2
\approx
d_{\mathrm{tax}}(u,v)$.
Node2Vec provides local smoothness, while MDS imposes global hierarchical separation. 
Figure~\ref{fig:representation_geometry}b visualizes the distinction between these two embedding spaces using dot-product similarity matrices. Node2Vec captures local taxonomic neighborhoods but can blur higher-level boundaries, whereas MDS produces sharper block structure aligned with genus- and family-level separation. 

\subsubsection{Genome Encoder and Alignment Objective}

Given the hyperspectral tensor $\mathrm{X} \in \mathbb{R}^{4^k \times 4^k \times C}$, we learn a genome encoder $f_{\theta}$ that maps $\mathrm{X}$ to a $d$-dimensional latent representation:
$\mathbf{h}(\mathrm{G}){=}f_{\theta}(\mathrm{X}).$ 
Because $\mathrm{X}$ is a multichannel image-like representation of motif-transition structure, we use a convolutional neural network (CNN) encoder. The CNN extracts local transition patterns and composes them into higher-level features through stacked convolution and pooling operations. This provides a natural inductive bias for hyperspectral genome tensors while avoiding the sequence-length scaling challenges associated with recurrent or transformer-based whole-genome encoders. 
For a genome $\mathrm{G}$ belonging to species $v$, the encoder is trained to align $\mathbf{h}(\mathrm{G})$ with the taxonomy-derived embedding of that species: 
$\mathcal{L}_{\mathrm{align}}
=
\left|
\mathbf{h}(\mathrm{G}) -
\mathbf{e}_{v}
\right|_2^2$.
Training proceeds in two stages. In Stage 1, $\mathbf{e}_{v}=\mathbf{e}^{\mathrm{n2v}}_v$, which initializes the encoder using locally smooth taxonomy neighborhoods. In Stage 2, $\mathbf{e}_{v}=\mathbf{e}^{\mathrm{mds}}_v$, which fine-tunes the encoder toward globally consistent hierarchical separation. 
This Node2Vec-to-MDS alignment approximates the composite geometry in Eq.~\eqref{eq:tax_geometry} while remaining trainable using genome-level supervision. 
We use a regression objective rather than a contrastive objective because microbial genome datasets are often sparse and imbalanced across species. Contrastive learning typically requires batches with multiple representatives per class and carefully sampled negatives, which may not be available for rare or newly sequenced organisms. 

\subsubsection{Inference} 
At inference time, a query genome $\mathrm{G}_{q}$ is converted into its hyperspectral tensor $\mathrm{X}_{q}$ and embedded as $\mathbf{h}(\mathrm{G}_{q}) = f_{\theta}(\mathrm{X}_{q})$.
Taxonomic prediction is performed by nearest-neighbor matching between $\mathbf{h}(\mathrm{G}_{q})$ and candidate taxon embeddings in the taxonomy-aligned latent space to support zero-shot and generalized zero-shot taxonomic prediction.

\noindent\textbf{Zero-shot inference.}
For unseen species, training genomes are unavailable, but the MDS-derived taxonomy embedding $\mathbf{e}^{\mathrm{mds}}_v$ can be computed from the taxonomy graph. The predicted species is 
$\hat{v}=\arg\min_{v \in \mathcal{S}_{\mathrm{unseen}}}
\left|
\mathbf{h}(\mathrm{G}_{q}) -
\mathbf{e}^{\mathrm{mds}}_v
\right|_2$. 
This enables species-level prediction without observing any genome from the target species during training.

\noindent\textbf{Generalized zero-shot inference.}
In the GZSL setting, the candidate set contains both seen and unseen species:
\begin{equation}
\hat{v}=\arg\min_{v \in \mathcal{S}_{\mathrm{seen}} \cup \mathcal{S}_{\mathrm{unseen}}}
\left|
\mathbf{h}(\mathrm{G}_{q}) -
\mathbf{e}^{\mathrm{mds}}_v
\right|_2.
\end{equation}
GZSL jointly evaluates seen and unseen species, requiring predictions over the combined candidate set. 

\noindent\textbf{Cascaded genus-to-species inference.}
We also evaluate a hierarchical inference strategy. First, the query is matched to genus-level embeddings, obtained by averaging the embeddings of species within each genus. Species prediction is then restricted to members of the top-ranked genus or genera, and the final label is assigned by nearest-neighbor matching within this reduced candidate set. This cascade reduces cross-genus confusion and captures the coarse-to-fine taxonomic structure.

\begin{figure}[t]
    \centering
    \includegraphics[width=0.8\linewidth]{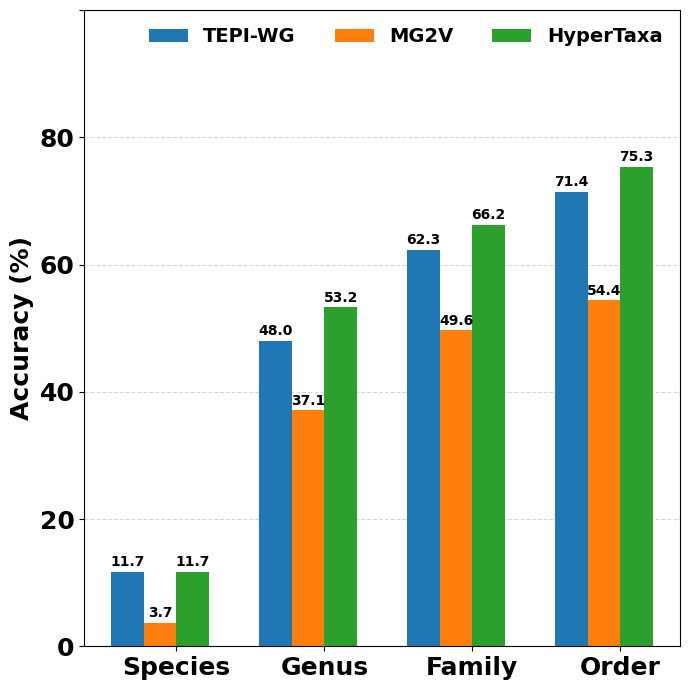}
    \caption{\textbf{Zero-shot Top-1 accuracy} on the TEPI benchmark.}
    \label{fig:rep}
\end{figure}

\subsection{Experimental Setup}

\textbf{Datasets.}
We evaluated \textsc{HyperTaxa} in three settings. First, we used the 97-species TEPI benchmark~\cite{aakur2024tepi} under its original zero-shot protocol. Second, we constructed a large-scale dataset from NCBI RefSeq release 216 ``Complete Genome'' assemblies~\cite{Goldfarb2025}, treating each assembly as one sample and concatenating chromosomes and plasmids in RefSeq order. For ZSL/GZSL, we trained on 1{,}110 species and evaluated on 219 held-out species whose families were represented during training. Held-out species had no genome examples in training; their class embeddings were derived only from the taxonomy graph. Sparse training classes were balanced with noisy synthetic hyperspectral images, with no synthetic data used for evaluation. Third, for within-species generalization, we selected species with at least two complete genomes, using one for training and the remainder for testing, yielding 736 training and 414 test images across 184 species.

\textbf{Implementation and Evaluation.}
We report Top-1 and Top-5 accuracy at species, genus, family, and order ranks. ZSL includes only held-out species, while GZSL evaluates seen and unseen species jointly. Cascaded inference predicts genus first and then restricts species prediction to that genus. 
All variants used the same splits and alignment protocol. We compared the proposed CNN with ResNet-50~\cite{he2016deep} and ViT-B/16~\cite{dosovitskiy2020image}, and evaluated three- versus five-channel hyperspectral inputs and single-stage Node2Vec- or MDS-aligned variants. All encoders produced $d{=}128$ embeddings and were trained with SGD (learning rate $10^{-3}$, batch size 32). Unless otherwise specified, we used $k{=}6$, yielding $4096\times4096\times C$ tensors, with $C{=}5$ strides $\{5,10,15,20,25\}$ and $\alpha_n^{(s)}{=}1$; three-channel inputs were used only for ablation. 

\begin{figure}
\centering
\begin{tabular}{c}
\textbf{a. Zero-Shot (ZSL) Performance} \\
\includegraphics[width=0.95\linewidth]{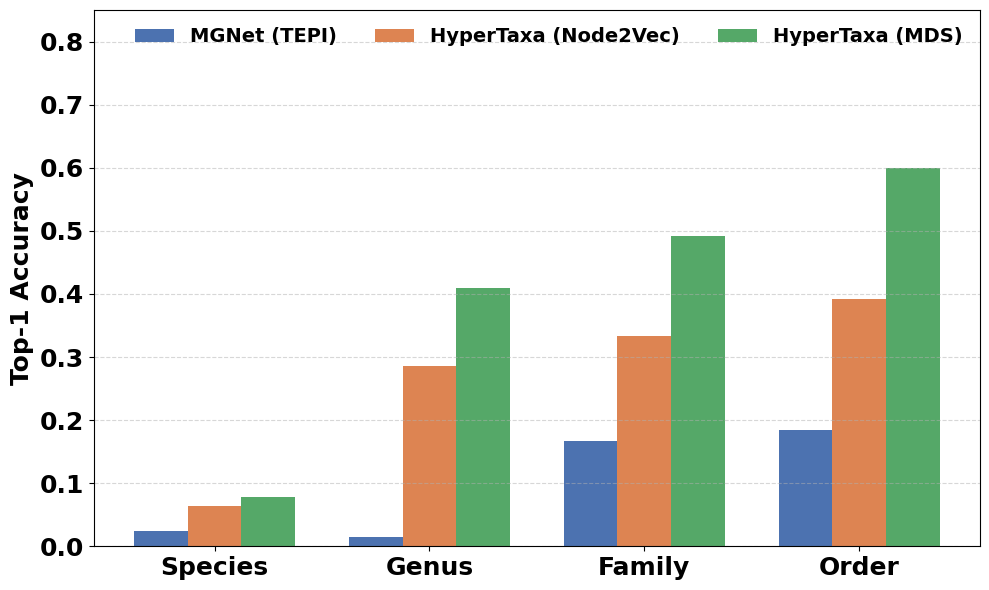} \\
\textbf{b. Generalized Zero-Shot (GZSL) Performance}\\
\includegraphics[width=0.95\linewidth]{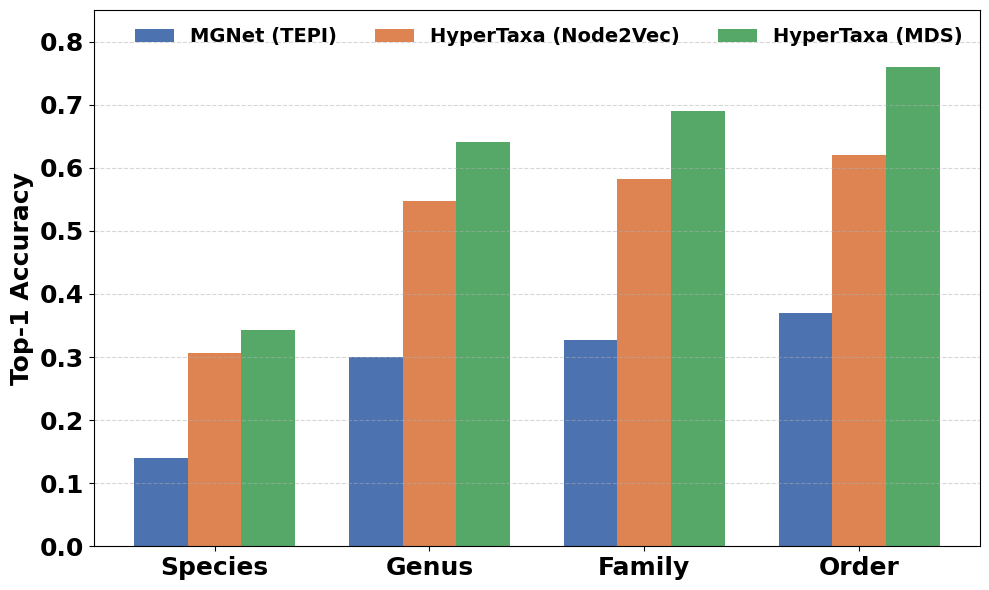} \\
\end{tabular}
\caption{\textbf{ZSL and GZSL performance.}
\textbf{a.} \textsc{HyperTaxa} outperforms MGNet/TEPI across all ranks in ZSL.
\textbf{b.} In GZSL, MDS alignment yields the strongest performance, indicating improved hierarchical generalization.}
\label{fig:zsl_gzsl}
\end{figure}

\begin{figure*}[t]
\centering
\begin{tabular}{ccc}
\textbf{a. Cascaded Prediction} &
\textbf{b. Hyperspectral Channel Ablation} &
\textbf{c. Genome Encoder Ablation} \\
\includegraphics[width=0.31\linewidth]{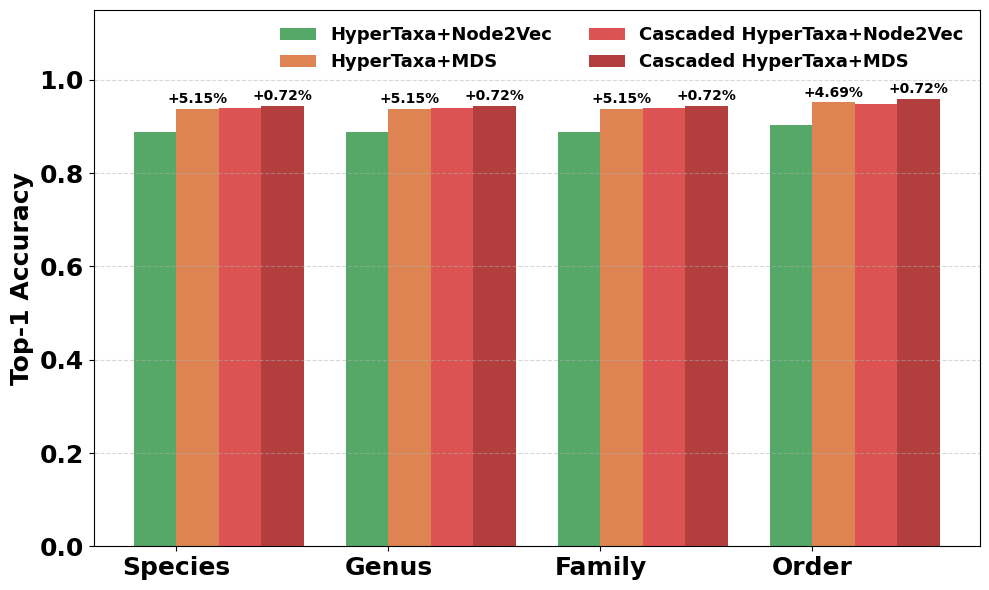} & 
\includegraphics[width=0.31\linewidth]{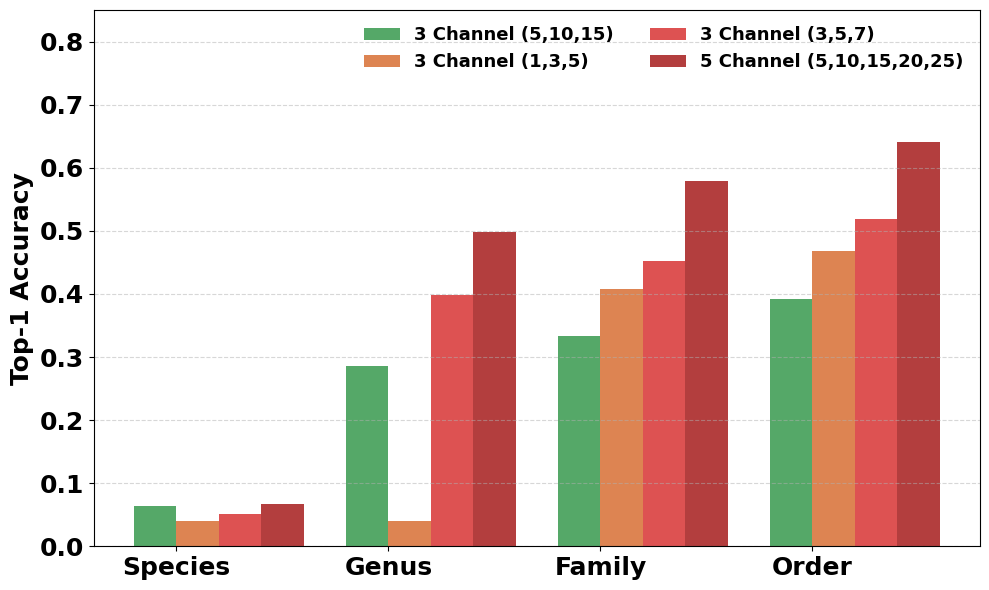} &
\includegraphics[width=0.31\linewidth]{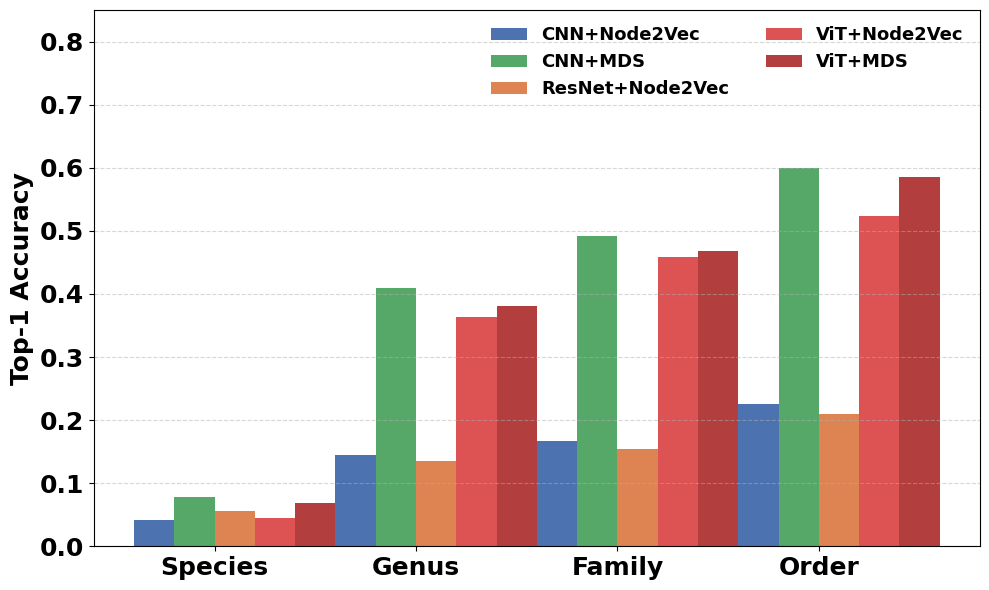}\\
\end{tabular}
\caption{\textbf{Hierarchical inference and ablations.}
\textbf{a.} Cascaded genus$\rightarrow$species prediction improves accuracy.
\textbf{b.} Increasing the number and range of hyperspectral $k$-mer transition channels improves zero-shot generalization.
\textbf{c.} CNN+MDS provides the strongest zero-shot performance among encoder and alignment variants.}
\label{fig:case3_ablation}
\end{figure*}

\section{Results}

\subsection{Multiscale Representation Improves Zero-Shot Prediction}
We first evaluated whether the proposed hyperspectral genome representation improves zero-shot taxonomic inference relative to prior genome pseudo-image and transformer-graph baselines. On TEPI's 97-species benchmark~\cite{aakur2024tepi}, \textsc{HyperTaxa} was trained and evaluated under the same held-out species protocol used by TEPI. As shown in Figure~\ref{fig:representation_geometry}, the TEPI pseudo-image representation produces sparse position-derived patterns, whereas the \textsc{HyperTaxa} image exhibits structured multiscale $k$-mer transition motifs. These differences translate into improved zero-shot accuracy across taxonomic ranks. \textsc{HyperTaxa} matches TEPI-WG at the species level with 11.7\% Top-1 accuracy, while substantially improving genus, family, and order accuracy to 53.2\%, 66.2\%, and 75.3\%, respectively. These gains indicate that multiscale transition fields provide a stronger substrate for learning taxonomically meaningful genome representations than pseudo-image or transformer-graph alternatives. 
We also examined the taxonomy-derived embedding spaces used to guide alignment. Node2Vec embeddings preserve local graph neighborhoods, but their similarity structure shows diffuse higher-level separation. In contrast, MDS embeddings derived from taxonomy distances produce sharper block structure aligned with genus- and family-level hierarchy. This supports the two-stage alignment strategy used by \textsc{HyperTaxa}: Node2Vec provides locally smooth initialization, while MDS imposes global taxonomic geometry.

\subsection{Zero-Shot and Generalized Zero-Shot Evaluation}

We next evaluated \textsc{HyperTaxa} on a larger disjoint-species split with 1{,}110 training and 219 unseen test species. Because held-out species share higher-level taxa with training species, this setting evaluates generalization to unseen species and their hierarchical localization within known taxonomic lineages. 
Figure~\ref{fig:zsl_gzsl}a shows that \textsc{HyperTaxa} outperforms the MGNet/TEPI pseudo-image baseline across all ranks in the strict ZSL setting. At the species level, accuracy improves from approximately 2--3\% for MGNet to 7--8\% for \textsc{HyperTaxa}. The gains are larger at higher ranks, with the MDS-aligned model reaching 49.0\% family-level and 37.0\% order-level Top-1 accuracy. Top-5 results follow the same trend, with \textsc{HyperTaxa} placing the correct species in the shortlist for more than half of unseen-species queries. 
In the GZSL setting, where seen and unseen species are evaluated jointly, \textsc{HyperTaxa} again improves over MGNet/TEPI (Figure~\ref{fig:zsl_gzsl}b). The MDS-aligned model achieves approximately 34--35\% species-level accuracy and 64--70\% accuracy at genus, family, and order ranks.

\subsection{Hierarchical Inference and Ablation Studies}

We evaluated whether explicit hierarchical inference improves prediction by applying a cascaded genus$\rightarrow$species strategy. As shown in Figure~\ref{fig:case3_ablation}a, restricting species-level prediction to candidates within the predicted genus improves performance over direct prediction. For the MDS-aligned model, species-level accuracy increases by approximately 5.1\%, with consistent gains across higher ranks. This suggests that the learned embedding space preserves useful coarse-to-fine taxonomic structure and that genus-level filtering reduces cross-genus confusion. 
Ablation studies further support the main design choices. Increasing the number of hyperspectral channels improves zero-shot performance across taxonomic ranks (Figure~\ref{fig:case3_ablation}b). The five-channel configuration using strides ${5,10,15,20,25}$ achieves the best overall performance, indicating that broader multiscale transition structure is important for open-set generalization. Encoder ablations show that the lightweight CNN paired with MDS alignment performs best in the ZSL setting (Figure~\ref{fig:case3_ablation}c). Deeper ResNet and ViT backbones do not improve zero-shot performance, suggesting that strong local inductive bias and taxonomy-aligned supervision are more important than model capacity for this representation.

\subsubsection{Qualitative Discussion}

The results support taxonomy-aware representation learning for open-set microbial taxonomic inference. \textsc{HyperTaxa} combines multiscale $k$-mer transition representations with Node2Vec-to-MDS alignment, enabling inference over species absent from genome-level training but represented in the taxonomy graph. Gains are strongest at genus, family, and order ranks, while cascaded genus-to-species inference reduces cross-genus confusion and ablations indicate that representation and embedding geometry matter more than model capacity. \textsc{HyperTaxa} is therefore complementary to tools such as BLAST, Kraken, GTDB-Tk, ANI/AAI, and TYGS rather than a replacement for deterministic sequence comparison. The current ZSL setting assumes unseen species are present in the taxonomy graph and share higher-level neighborhoods with training taxa, so truly unknown lineages may fall outside the learned space; experiments also focus on complete genomes rather than fragmented assemblies, MAGs, or raw-read mixtures.

\section{Conclusion and Future Work}

We introduced \textsc{HyperTaxa}, a taxonomy-aligned representation learning framework for open-set microbial taxonomic inference. By encoding genomes as multiscale hyperspectral $k$-mer transition fields and aligning them to a structured taxonomy-derived embedding space, \textsc{HyperTaxa} supports zero-shot, generalized zero-shot, and cascaded hierarchical inference. Results across benchmark, large-scale disjoint-species, and ablation studies show that taxonomy-aware genome representations improve inference for species excluded from genome-level training, particularly at higher taxonomic ranks. 
Future work will extend \textsc{HyperTaxa} to stricter reference-masking and phylogenetic holdout settings, fragmented genomes, and raw-read metagenomic data. We also plan to evaluate diagnostic-style candidate retrieval and hybrid workflows with established profilers and genome-based classification tools, including their effects on search-space reduction, runtime, and failure modes.

{
    \small
    \bibliographystyle{ieeetr}
    \bibliography{sample}
}
\end{document}